\documentclass{svproc}
\usepackage{graphicx}%
\usepackage{multirow}%
\usepackage{amsmath,amssymb,amsfonts}%
\usepackage{mathrsfs}%
\usepackage[title]{appendix}%
\usepackage{xcolor}%
\usepackage{textcomp}%
\usepackage{manyfoot}%
\usepackage{booktabs}%
\usepackage{algorithm}%
\usepackage{algorithmicx}%
\usepackage{algpseudocode}%
\usepackage{listings}%
\usepackage{url}

\begin{document}
	\mainmatter              
	\title{Enhancing Distance-Based Graph Autoencoders with Structural Penalties for Dynamic Graph Embedding}
	\titlerunning{GA with Structural Penalties for Dynamic Graph Embeddings}  
	%
	\author{Aleksandar Tom\v{c}i\'{c} \and Milo\v{s} Savi\'{c} \and Milo\v{s} Radovanovi\'{c}}
	\authorrunning{A. Tom\v{c}i\'{c} et al.} 

	\institute{Department of Mathematics and Informatics, Faculty of Sciences, \\ University of Novi Sad, Serbia\\
		Trg Dositeja Obradovi\'{c}a 4, 21000 Novi Sad, Serbia\\
		\email{\{atomic, svc, radacha\}@dmi.uns.ac.rs}}
	
	\maketitle              
	
	\begin{abstract}
		Graph autoencoders (GAEs) are widely used for learning representations of dynamic graphs. However, their optimization objectives typically do not take structural heterogeneity across nodes into account. We propose three distance-based GAE variants that incorporate structural penalties into the reconstruction loss. All variants share a two-layer Graph Convolutional Network encoder and a Euclidean-distance decoder trained with distance-based reconstruction objectives.
		We extend a sparsity-corrected reconstruction loss with two node-level weighting schemes that multiplicatively re-weight per-pair reconstruction errors: (i) a hub penalty based on degree centrality, and (ii) a penalty based on Natural Community Local Intrinsic Dimensionality (NC-LID). The paper is motivated by prior evidence linking high NC-LID to reduced embedding quality. The proposed methods are designed to emphasize reconstruction errors for structurally ambiguous nodes.
		Experiments on nine dynamic graph data sets show that NC-LID-based weighting yields the best reconstruction performance on six of them, compared with the unweighted baseline and the hub-aware variant. These findings highlight NC-LID as a useful structural signal for enhancing distance-based graph autoencoders in dynamic settings.
	\end{abstract}
	\section{Introduction}\label{sec:intro}

	Many real-world complex systems are inherently dynamic. The networks that represent them evolve
	over time through the addition or removal of nodes and edges~\cite{holme2012temporal}. Learning
	useful representations of such evolving structures is a core challenge in machine learning on
	graphs~\cite{barros2021survey}. Graph embedding methods offer
	a task-agnostic solution, as they map nodes to low dimensional vectors that preserve structural properties, and can subsequently be used for a wide range of downstream tasks including link prediction, node classification, and anomaly detection~\cite{khoshraftar2024survey}.
	
	Among dynamic graph embedding approaches, two main families exist. Methods based on random walks~\cite{mahdavi2018dynnode2vec,nguyen2018continuous,pandhre2018stwalk} generate
	node contexts by traversing the graph and learn embeddings from the resulting sequences, following
	the paradigm of static walk-based methods such as node2vec~\cite{grover2016node2vec} and
	DeepWalk~\cite{perozzi2014deepwalk}. Methods based on autoencoders~\cite{barros2021survey} encode the graph structure directly through a neural network and reconstruct it from a
	low-dimensional bottleneck. While walk-based methods have received more research attention,
	autoencoder-based methods remain comparatively underexplored in terms of structural signals that
	could improve their training objectives.
	
	Recent lines of work are particularly relevant here. First, it has been shown
	that hub-awareness, incorporating node degree centrality into walk transition probabilities,
	substantially improves dynamic graph embedding~\cite{tomvcic2025dynamic}.
	Second, NC-LID, a graph-adapted measure of local intrinsic dimensionality~\cite{savic2021local,savic2023local},
	has been shown to negatively correlate with node embedding quality in dynamic graphs. Nodes with
	high NC-LID (complex, irregularly-shaped natural communities) tend to be systematically
	underrepresented in walk-based dynamic embeddings~\cite{knevzevic2024local}. Both findings point to
	structural properties that the standard training objectives of autoencoder-based methods do not take into account.
	
	This paper investigates whether these structural signals can be incorporated directly into the
	loss function of a distance-based graph autoencoder. We make the following contributions:
	
	\begin{itemize}
		\item We design a \textbf{Euclidean distance decoder} for dynamic graph autoencoders that
		aligns the training objective with the Euclidean distance-based graph reconstruction criterion used in evaluation, motivated by the geometric mismatch of the standard inner product decoder.
		\item We introduce two \textbf{structurally penalized loss functions}, i.e., per-pair re-weightings of the reconstruction loss, one based on the degree outer product (hub penalty) and one based on the NC-LID outer product (LID penalty), each amplifying reconstruction errors for structurally significant node pairs.
		\item We provide a systematic empirical evaluation on nine dynamic networks,
		showing that the LID-penalized variant achieves better reconstruction $F_1$ score in six of
		nine datasets while adding negligible computational overhead through a precomputation
		strategy for NC-LID scores.
	\end{itemize}

	\section{Related Work}\label{sec:related}
	
	Surveys by Barros et al.~\cite{barros2021survey} and Khoshraftar and An~\cite{khoshraftar2024survey}
	provide comprehensive overviews of dynamic graph embedding methods. Among walk-based methods,
	Dynnode2vec~\cite{mahdavi2018dynnode2vec} initializes embeddings via node2vec~\cite{grover2016node2vec}
	and incrementally updates only nodes whose ego networks change; CTDNE~\cite{nguyen2018continuous}
	enforces temporal ordering within walks, and STWalk~\cite{pandhre2018stwalk} combines spatial and
	historical walks. Autoencoder-based methods are surveyed in~\cite{barros2021survey}, but remain
	comparatively underexplored in terms of structural inductive biases.
	
	In the static setting, hub-aware random walk methods~\cite{tomvcic2024hub} that adjust transition
	probabilities by node degree improve node classification performance; DeepHub~\cite{tomvcic2025dynamic}
	extends this to the dynamic setting, showing that inverse-degree-biased walks outperform Dynnode2vec
	on graph reconstruction while narrowing the hub/non-hub embedding quality gap.
	
	Recent work~\cite{savic2021local,savic2023local} introduced NC-LID, a local intrinsic dimensionality
	measure defined over natural communities~\cite{lancichinetti2009detecting}: low NC-LID indicates
	compact, well-defined communities, while high NC-LID indicates structurally ambiguous nodes spanning
	multiple structural directions. LID-elastic node2vec~\cite{savic2023local} improved both intrinsic
	embedding quality and downstream performance in the static setting, and~\cite{knevzevic2024local}
	adapted NC-LID to the dynamic setting, showing significant negative Spearman correlations between
	NC-LID and embedding $F_1$ under Dynnode2vec, establishing NC-LID as a reliable indicator of weakly
	embedded nodes.
	
	Kipf and Welling~\cite{kipf2016variational} introduced the GAE framework, combining GCN
	encoders~\cite{kipf2016semi} with inner product decoders. This decoder has become standard, but its
	implicit assumption, that co-directional vectors correspond to connected nodes, is inconsistent with
	Euclidean distance-based evaluation. Distance-based decoders have been explored for hyperbolic
	embeddings~\cite{nickel2017poincare} and link prediction~\cite{zhang2018link}, but not, to our
	knowledge, systematically applied to dynamic graph autoencoders with structurally informed losses.

	\section{Proposed Methods}\label{sec:method}
	
	\subsection{Background: Graph Autoencoders and GCNs}
	
	An autoencoder is a neural network trained to reconstruct its own input through a low-dimensional
	bottleneck. The encoder $f_\theta : \mathcal{X} \to \mathcal{Z}$ maps an input
	$x \in \mathcal{X}$ to a latent code $z \in \mathcal{Z}$, and the decoder
	$g_\phi : \mathcal{Z} \to \mathcal{X}$ attempts to recover $x$ from $z$:
	\begin{equation}
		\min_{\theta,\phi}\; \mathbb{E}_{x \sim p(x)}\!\left[
		\mathcal{L}\!\left(x,\, g_\phi(f_\theta(x))\right)
		\right].
		\label{eq:ae_objective}
	\end{equation}
	Since $|\mathcal{Z}| \ll |\mathcal{X}|$, the model must discard noise and retain only the most
	salient structure of the input.
	
	A Graph Autoencoder (GAE)~\cite{kipf2016variational} instantiates this with a GCN encoder~\cite{kipf2016semi}.
	At layer $\ell$, node representations are updated by:
	\begin{equation}
		\mathbf{H}^{(\ell)} = \sigma\!\left(\tilde{\mathbf{A}}\,\mathbf{H}^{(\ell-1)}\,\mathbf{W}^{(\ell)}\right),
		\label{eq:gcn}
	\end{equation}
	where $\tilde{\mathbf{A}} = \hat{\mathbf{D}}^{-1/2}(\mathbf{A}+\mathbf{I})\hat{\mathbf{D}}^{-1/2}$
	is the symmetrically normalized adjacency with self-loops and $\mathbf{W}^{(\ell)}$ is a learnable
	weight matrix. After $L$ layers, the embedding of node $v_i$ integrates the structure of all nodes
	within graph distance $L$, e.g., two layers encode both immediate and two-hop neighborhood structure.
	
	\subsection{Dynamic Graph Setting}
	
	Let $\mathcal{G} = \{G_1, \dots, G_T\}$ denote a discrete-time dynamic graph, where each
	snapshot $G_t = (V_t, E_t)$ is a static graph on $n_t = |V_t|$ nodes, and let
	$\mathcal{V} = \bigcup_t V_t$ be the global vocabulary of $N = |\mathcal{V}|$ distinct nodes.
	Since these datasets carry no semantic node attributes, each node is represented by a one-hot
	identity vector keyed by its global vocabulary index, giving feature matrix
	$\mathbf{X}^{(t)} \in \{0,1\}^{n_t \times N}$ with $X^{(t)}_{i,\texttt{idx}(v_i)} = 1$ elsewhere zero.
	This global vocabulary keeps the encoder input dimension consistent across snapshots without padding or masking, regardless of node dynamics.

	\textit{Scope.} We adopt a snapshot-wise formulation, a separate model is trained for each $G_t$ from random initialization, without parameter sharing or temporal smoothing, and evaluated on the reconstruction of that snapshot. The temporal dimension therefore enters through the evolving topology of each snapshot and through the per-snapshot NC-LID scores, which track how structural ambiguity of nodes changes over time. The model does not explicitly learn temporal dependencies, and embeddings of the same node at different times are not aligned and should not be compared directly. The proposed decoder and loss weighting are thus not specific to dynamic graphs, and can be combined with temporally coupled training schemes (e.g., warm-starting from the previous snapshot), which we leave to future work.
	
	\subsection{Euclidean Distance Decoder}
	
	Classical GAEs decode via the inner product $\hat{A}_{ij} = \sigma(\mathbf{z}_i^\top\mathbf{z}_j)$,
	assigning high scores to co-directional rather than geometrically close vectors. Since graph
	reconstruction evaluation connects the $|E_t|$ closest node pairs by Euclidean
	distance~\cite{tomvcic2025dynamic,knevzevic2024local}, training with an inner product decoder
	mismatches the objective and the evaluation metric.
	
	Our \textbf{EuclideanGAEModel} replaces the inner product decoder with one based on pairwise
	squared Euclidean distances. The encoder uses $\tanh$ in the hidden layer to bound embedding
	coordinates and prevent distance explosion:
	\begin{align}
		\mathbf{H} &= \tanh\!\left(\tilde{\mathbf{A}}\,\mathbf{X}\,\mathbf{W}^{(1)}\right), \label{eq:enc1}\\
		\mathbf{Z} &= \tilde{\mathbf{A}}\,\mathbf{H}\,\mathbf{W}^{(2)}, \label{eq:enc2}
	\end{align}
	with $\mathbf{W}^{(1)} \in \mathbb{R}^{N \times d_h}$, $\mathbf{W}^{(2)} \in \mathbb{R}^{d_h \times d}$,
	and hidden dimension $d_h = 2d$. The decoder forms the full pairwise distance matrix using the
	numerically stable expansion:
	\begin{equation}
		D_{ij} = \max\!\bigl(0,\;s_i + s_j - 2[\mathbf{Z}\mathbf{Z}^\top]_{ij}\bigr),
		\quad s_i = \textstyle\sum_k Z_{ik}^2,
		\label{eq:dist}
	\end{equation}
	and converts distances to adjacency logits via a learnable scalar bias $b$ (initialized to zero):
	\begin{equation}
		\hat{A}_{ij} = b - D_{ij}.
		\label{eq:logit}
	\end{equation}
	Node pairs with $D_{ij} < b$ receive positive logits (edge predicted); pairs with $D_{ij} > b$
	receive negative logits (no edge predicted). The bias adapts to the average connectivity of each
	snapshot. The computational cost of the decoder is $O(n^2 d)$, identical to the inner product
	decoder, since the dominant operation is $\mathbf{Z}\mathbf{Z}^\top$.
	
	\subsection{Sparsity-Corrected Baseline Loss}\label{sec:base_loss}
	
	Real-world networks are highly sparse, the edge density $|E_t|/n_t^2$ is below $1\%$ for most
	datasets studied here. A na\"{i}ve binary cross entropy (BCE) loss is dominated by the abundant
	negative pairs, biasing the model toward predicting no edges. Following standard practice, we
	up-weight positive pairs by the class imbalance ratio:
	
	\begin{equation}
		w^{+} = \frac{n^2 - |E|}{|E| + \varepsilon}, \qquad \varepsilon = 10^{-9},
		\label{eq:pos_weight}
	\end{equation}
	
	\noindent yielding the sparsity corrected baseline loss used by \textbf{Distance\_Basic} variant:
	
	\begin{equation}
		\mathcal{L}_{\mathrm{base}} = \frac{1}{n^2}\sum_{i,j} w_{ij}\;\ell_{\mathrm{BCE}}\!\left(\hat{A}_{ij},A_{ij}\right),
		\quad w_{ij} = \begin{cases} w^{+} & A_{ij}=1 \\ 1 & A_{ij}=0.\end{cases}
		\label{eq:base_loss}
	\end{equation}
	
	\noindent where $\ell_{\mathrm{BCE}}(\hat{a}, a) = -a\log\sigma(\hat{a}) -
	(1-a)\log(1-\sigma(\hat{a}))$ and $\sigma$ is the sigmoid function.
	
	\subsection{Hub-Penalty Variant}\label{sec:hub_loss}
	
	High-degree nodes (\emph{hubs}) occupy structurally critical positions~\cite{radovanovic2010hubs,tomvcic2024hub,tomvcic2025dynamic},
	acting as bridges between communities and appearing disproportionately in shortest paths, so
	incorrectly placing a hub propagates reconstruction error to all $\deg(v)$ incident pairs. The
	\textbf{Distance\_Hub} variant amplifies the reconstruction loss for hub-involved pairs in
	proportion to the product of their degrees:
	\begin{equation}
		\delta_{ij} = \deg(v_i)\cdot\deg(v_j),
		\qquad
		\tilde{\delta}_{ij} = \frac{\delta_{ij}}{\max_{k,l}\delta_{kl}+\varepsilon}.
		\label{eq:hub_delta}
	\end{equation}
	\begin{equation}
		\mathcal{L}_{\mathrm{hub}} = \frac{1}{n^2}\sum_{i,j}
		w_{ij}\;\ell_{\mathrm{BCE}}\!\left(\hat{A}_{ij},A_{ij}\right)\cdot
		\bigl(1 + \alpha_h\,\tilde{\delta}_{ij}\bigr),
		\label{eq:hub_loss}
	\end{equation}
	where $\alpha_h \geq 0$ controls penalty strength (set to $1.0$ in all experiments). The
	multiplier $(1+\alpha_h\,\tilde{\delta}_{ij})\in[1,1+\alpha_h]$ ensures the baseline loss is
	only amplified, never suppressed. Note that, despite the term \emph{penalty}, neither variant adds a regularization term to the objective, both re-weight the per-pair reconstruction loss.
	
	\subsection{LID-Penalty Variant}\label{sec:lid_loss}
	
	The \textbf{Distance\_LID} variant replaces the degree-based signal with NC-LID, a topological
	measure of structural ambiguity. Following work presented in~\cite{savic2023local}, the NC-LID of a
	node $v$ is defined over its natural community $S$ and the largest shortest-path distance $k$ from $v$ to any node in $S$. As in~\cite{knevzevic2024local,savic2023local}, the natural community is recovered by the local fitness-based algorithm of Lancichinetti et al.~\cite{lancichinetti2009detecting} with $v$ as the seed node, starting from $S=\{v\}$, the community is grown greedily by adding the neighboring node that most increases a fitness function comparing the internal and external degrees of $S$, while nodes whose removal increases the fitness are removed, the process stops when no neighboring node increases the fitness. NC-LID is then:
	
	\begin{equation}
		\mathrm{NC\text{-}LID}(v) = -\ln\!\left(\frac{|S|}{D(v,k)}\right),
		\label{eq:nclid}
	\end{equation}
	
	\noindent where $D(v,k)$ is the number of nodes at shortest-path distance at most $k$ from $v$. A value
	of zero indicates a perfectly compact community; higher values indicate that the shortest-path
	distance struggles to separate $S$ from the rest of the graph, reflecting structural ambiguity.
	As shown in~\cite{knevzevic2024local}, NC-LID correlates negatively with embedding
	quality in dynamic graphs for the majority of datasets.
	
	Amplifying the reconstruction penalty for boundary nodes (high NC-LID on both endpoints) forces
	the encoder to resolve their positions precisely rather than collapsing them into the interior of
	the nearest dominant community. The loss is:
	
	\begin{equation}
		\lambda_{ij} = \mathrm{NC\text{-}LID}(v_i)\cdot\mathrm{NC\text{-}LID}(v_j),
		\qquad
		\tilde{\lambda}_{ij} = \frac{\lambda_{ij}}{\max_{k,l}\lambda_{kl}+\varepsilon},
		\label{eq:lid_outer}
	\end{equation}
	
	\begin{equation}
		\mathcal{L}_{\mathrm{lid}} = \frac{1}{n^2}\sum_{i,j}
		w_{ij}\;\ell_{\mathrm{BCE}}\!\left(\hat{A}_{ij},A_{ij}\right)\cdot
		\underbrace{\bigl(1 + \alpha_\ell\,\tilde{\lambda}_{ij}\bigr)}_{\text{LID multiplier}},
		\label{eq:lid_loss}
	\end{equation}
	
	\noindent with $\alpha_\ell = 1.0$ throughout. The LID multiplier lies in $[1,1+\alpha_\ell]$, matching the hub variant's guarantee
	that the baseline loss is never suppressed.
	
	\subsection{Computational Cost}\label{sec:cost}

	All three variants share the encoder and decoder, so their per-epoch cost is identical up to the loss, the two sparse GCN propagations cost $O(|E_t|\,d_h)$, the dense feature transformations $O(n_t N d_h + n_t d_h d)$, and the decoder $O(n_t^2 d)$, which dominates for the snapshot sizes considered here. \textbf{Distance\_Basic} adds an $O(n_t^2)$ weighted BCE. \textbf{Distance\_Hub} and \textbf{Distance\_LID} each add one $O(n_t^2)$ outer product and element-wise multiplication; degrees are read directly from the snapshot. The only additional cost of \textbf{Distance\_LID} is NC-LID estimation, which requires a greedy natural community expansion and a bounded breadth-first search per node. Since NC-LID depends only on the topology of $G_t$ and not on the embeddings, it is computed once per snapshot before training, cached, and reused across all epochs and embedding dimensions. Nothing is lost by this caching, recomputing NC-LID at every epoch would return identical values while multiplying the estimation cost by $T_e$, whereas evolving structural ambiguity over time is already captured because NC-LID is recomputed for every snapshot. With caching, the per-epoch training cost of \textbf{Distance\_LID} stays within $\pm10\%$ of the baseline across all datasets.

	\section{Experimental Setup}\label{sec:experiments}

	The proposed methods are evaluated on nine publicly available temporal networks from SNAP\footnote{\url{https://snap.stanford.edu/data/\#temporal}} and Network Repository,\footnote{\url{https://networkrepository.com/dynamic.php}} summarized in Table~\ref{tab:datasets} (institutional email, physical proximity, and online social messaging networks; 54--1{,}394 nodes per snapshot, 2--18 snapshots).
	
	\begin{table}[t]
		\centering
		\caption{Experimental datasets. Max nodes and max edges refer to the largest individual snapshot.}\label{tab:datasets}%
		\setlength{\tabcolsep}{4pt}
		\small
		\begin{tabular}{llrrr}
			\toprule
			\textbf{Dataset} & \textbf{Res.} & \textbf{Snaps.} & \textbf{Max nodes} & \textbf{Max edges} \\
			\midrule
			radoslaw-email                   & month    &  9 &  151 & 1{,}675 \\
			ia-hospital-ward-proximity-attr  & day      &  5 &   54 &   492   \\
			ia-contacts\_hypertext2009       & day      &  3 &  102 & 1{,}062 \\
			ia-enron-employees               & 3-months & 13 &  140 &   823   \\
			ia-primary-school-proximity-attr & day      &  2 &  241 & 5{,}923 \\
			fb-forum                         & month    &  6 &  815 & 5{,}838 \\
			email-Eu-core-temporal           & month    & 18 &  762 & 4{,}518 \\
			CollegeMsg                       & month    &  7 & 1{,}371 & 6{,}865 \\
			fb-messages                      & month    &  7 & 1{,}394 & 9{,}425 \\
			\bottomrule
		\end{tabular}
	\end{table}
	
	Effective graph embeddings should allow reconstruction of the original graph: computing the
	Euclidean distance between every pair of embedding vectors and connecting the closest $|E|$ pairs,
	where $|E|$ is the number of edges in the original graph. Let $n$ denote an arbitrary node and $C$
	the number of correctly reconstructed links incident to $n$. We use:
	\begin{itemize}
		\item Precision -- $C$ divided by the number of links $n$ has in the reconstructed graph.
		\item Recall -- $C$ divided by the number of links $n$ has in the original graph.
		\item $F_1$ score -- The harmonic mean of precision and recall.
	\end{itemize}
	Higher values of precision, recall, and $F_1$ indicate fewer link reconstruction errors for node $n$. At the graph level, precision, recall, and $F_1$ scores can be obtained by macro-averaging over all nodes.
	We report mean $F_1$ averaged across all temporal snapshots of each dataset.
	
	Five embedding dimensions are evaluated: $d \in \{10, 25, 50, 100, 200\}$ with hidden dimension
	$d_h = 2d$. All models are trained for $T_e = 300$ epochs with Adam at $\eta = 0.01$.
	Each model dimension combination is trained from random initialization without parameter sharing
	across snapshots. Penalty strengths are fixed at $\alpha_h = \alpha_\ell = 1.0$ for all
	penalized variants.

	\section{Results and Discussion}\label{sec:results}
	
	\subsection{Per-Dataset Reconstruction $F_1$}
	
	Tables~\ref{tab:res1}-\ref{tab:res3} report mean reconstruction $F_1$ for all nine datasets across
	the five embedding dimensions. Bold entries mark the best model for each dimension-dataset pair.
	The symbol $\star$ marks the single best configuration per dataset.
	
	Table~\ref{tab:summary} condenses the results to the single best $F_1$ per model per dataset.
	
	\begin{table}[t]
	\centering
	\begin{minipage}[t]{0.485\textwidth}
	\centering
		
		\caption{Mean reconstruction $F_1$ for \textit{radoslaw-email} (top) and
			\textit{ia-hospital-ward-proximity-attr} (bottom).}\label{tab:res1}%
		\setlength{\tabcolsep}{4pt}
		\footnotesize
		\begin{tabular}{llll}
			\toprule
			$d$ & \textsc{Basic} & \textsc{Hub} & \textsc{LID} \\
			\midrule
			\multicolumn{4}{l}{\textit{radoslaw-email}} \\
			10  & 0.4597 & 0.4565 & \textbf{0.4777} \\
			25  & 0.5567 & 0.5408 & \textbf{0.5633} \\
			50  & 0.5917 & 0.5763 & \textbf{0.6016} \\
			100 & 0.6099 & 0.6000 & $\mathbf{0.6229}^{\star}$ \\
			200 & 0.6045 & 0.5933 & \textbf{0.6124} \\
			\midrule
			\multicolumn{4}{l}{\textit{ia-hospital-ward-proximity-attr}} \\
			10  & 0.6819          & 0.6893 & \textbf{0.6932} \\
			25  & 0.7729          & 0.7628 & \textbf{0.7792} \\
			50  & \textbf{0.8166} & 0.7920 & 0.7962 \\
			100 & \textbf{0.8155} & 0.7968 & 0.8132 \\
			200 & 0.8207          & 0.7891 & $\mathbf{0.8222}^{\star}$ \\
			\bottomrule
		\end{tabular}
		\end{minipage}\hfill
	\begin{minipage}[t]{0.485\textwidth}
	\centering
		
		\caption{Mean reconstruction $F_1$ for \textit{ia-contacts\_hypertext2009} (top),
			\textit{ia-enron-employees} (middle), and \textit{ia-primary-school-proximity-attr} (bottom).}\label{tab:res2}%
		\setlength{\tabcolsep}{4pt}
		\footnotesize
		\begin{tabular}{llll}
			\toprule
			$d$ & \textsc{Basic} & \textsc{Hub} & \textsc{LID} \\
			\midrule
			\multicolumn{4}{l}{\textit{ia-contacts\_hypertext2009}} \\
			10  & \textbf{0.5273} & 0.5120 & 0.5179 \\
			25  & 0.6373          & 0.6335 & \textbf{0.6417} \\
			50  & \textbf{0.6892} & 0.6727 & 0.6873 \\
			100 & $\mathbf{0.7290}^{\star}$ & 0.7197 & 0.7214 \\
			200 & \textbf{0.7136} & 0.7026 & 0.7130 \\
			\midrule
			\multicolumn{4}{l}{\textit{ia-enron-employees}} \\
			10  & \textbf{0.7359} & 0.7247 & 0.7238 \\
			25  & 0.7942          & 0.7672 & \textbf{0.7972} \\
			50  & 0.8033          & 0.7938 & \textbf{0.8083} \\
			100 & 0.8098          & 0.7999 & $\mathbf{0.8140}^{\star}$ \\
			200 & \textbf{0.8084} & 0.7945 & 0.8069 \\
			\midrule
			\multicolumn{4}{l}{\textit{ia-primary-school-proximity-attr}} \\
			10  & 0.7571          & \textbf{0.7588} & 0.7490 \\
			25  & 0.7644          & \textbf{0.7651} & 0.7635 \\
			50  & \textbf{0.7684} & 0.7669          & 0.7636 \\
			100 & 0.7694          & 0.7686          & \textbf{0.7706} \\
			200 & $\mathbf{0.7718}^{\star}$ & 0.7706 & 0.7698 \\
			\bottomrule
		\end{tabular}
		\end{minipage}
	\end{table}
	
	\begin{table}[t]
		\centering
		\caption{Mean reconstruction $F_1$ for \textit{fb-forum}, \textit{email-Eu-core-temporal},
			\textit{CollegeMsg}, and \textit{fb-messages}.}\label{tab:res3}%
		\setlength{\tabcolsep}{5pt}
		\small
		\begin{tabular}{llll}
			\toprule
			$d$ & \textsc{Basic} & \textsc{Hub} & \textsc{LID} \\
			\midrule
			\multicolumn{4}{l}{\textit{fb-forum}} \\
			10  & 0.3754 & 0.3562 & \textbf{0.4098} \\
			25  & 0.5797 & 0.5708 & \textbf{0.5925} \\
			50  & 0.6374 & 0.6277 & \textbf{0.6431} \\
			100 & 0.6522 & 0.6439 & \textbf{0.6622} \\
			200 & 0.6638 & 0.6600 & $\mathbf{0.6681}^{\star}$ \\
			\midrule
			\multicolumn{4}{l}{\textit{email-Eu-core-temporal}} \\
			10  & \textbf{0.3462} & 0.3387 & 0.3436 \\
			25  & 0.4364          & 0.4336 & \textbf{0.4415} \\
			50  & 0.4654          & 0.4598 & \textbf{0.4663} \\
			100 & 0.4773          & 0.4680 & \textbf{0.4789} \\
			200 & 0.4912          & 0.4767 & $\mathbf{0.4914}^{\star}$ \\
			\midrule
			\multicolumn{4}{l}{\textit{CollegeMsg}} \\
			10  & 0.2062          & 0.2231          & \textbf{0.2242} \\
			25  & 0.3476          & \textbf{0.3785} & 0.3736 \\
			50  & 0.4350          & 0.4261          & \textbf{0.4375} \\
			100 & $\mathbf{0.4641}^{\star}$ & 0.4238 & 0.4482 \\
			200 & \textbf{0.3758} & 0.3604          & 0.3613 \\
			\midrule
			\multicolumn{4}{l}{\textit{fb-messages}} \\
			10  & 0.1840          & \textbf{0.2033} & 0.2030 \\
			25  & \textbf{0.3648} & 0.3397          & 0.3377 \\
			50  & \textbf{0.4213} & 0.4132          & 0.4147 \\
			100 & 0.4320          & 0.4392          & $\mathbf{0.4471}^{\star}$ \\
			200 & 0.3594          & 0.3516          & \textbf{0.3818} \\
			\bottomrule
		\end{tabular}
	\end{table}

	\begin{table}[t]
		\centering
		\caption{Best mean reconstruction $F_1$ per dataset and model. $d^{\star}$ is the optimal dimension.
			Bold marks the winner per dataset. \emph{Win count} tallies best $F_1$ victories across all 9 datasets.}\label{tab:summary}%
		\setlength{\tabcolsep}{4pt}
		\small
		\begin{tabular}{lllllll}
			\toprule
			& \multicolumn{2}{c}{\textsc{Basic}} & \multicolumn{2}{c}{\textsc{Hub}}
			& \multicolumn{2}{c}{\textsc{LID}} \\
			\cmidrule(lr){2-3}\cmidrule(lr){4-5}\cmidrule(lr){6-7}
			\textbf{Dataset} & $F_1$ & $d^{\star}$ & $F_1$ & $d^{\star}$ & $F_1$ & $d^{\star}$ \\
			\midrule
			radoslaw-email     & 0.6099 & 100 & 0.6000 & 100 & \textbf{0.6229} & 100 \\
			hospital-ward      & 0.8207 & 200 & 0.7968 & 100 & \textbf{0.8222} & 200 \\
			hypertext2009      & \textbf{0.7290} & 100 & 0.7197 & 100 & 0.7214 & 100 \\
			enron-employees    & 0.8098 & 100 & 0.7999 & 100 & \textbf{0.8140} & 100 \\
			primary-school     & \textbf{0.7718} & 200 & 0.7706 & 200 & 0.7706 & 100 \\
			fb-forum           & 0.6638 & 200 & 0.6600 & 200 & \textbf{0.6681} & 200 \\
			email-Eu-core      & 0.4912 & 200 & 0.4767 & 200 & \textbf{0.4914} & 200 \\
			CollegeMsg         & \textbf{0.4641} & 100 & 0.4261 & 50 & 0.4482 & 100 \\
			fb-messages        & 0.4320 & 100 & 0.4392 & 100 & \textbf{0.4471} & 100 \\
			\midrule
			\textit{Win count} & 3 & & 0 & & \textbf{6} & \\
			\bottomrule
		\end{tabular}
	\end{table}

	\subsection{Discussion}
	
	Reconstruction $F_1$ increases consistently as $d$ grows from 10 to 100 across all models and
	datasets, then plateaus or declines at $d=200$, consistent with a capacity-generalisation
	trade-off: at $d=10$ the bottleneck is too tight for structurally distinct nodes, while at
	$d=200$ the encoder ($N\times 2d$ parameters) begins to memorize snapshot-specific pairs rather
	than learning generalisable codes. The best model reaches its optimum at $d=100$ on five of nine datasets; on the remaining four (hospital-ward, primary-school, fb-forum, email-Eu-core) $d=200$ is best, but by at most $1.3$~pp over $d=100$.
	
	\textbf{Distance\_LID} wins on six of nine datasets and records the top score in 26 of 45
	dimension-dataset configurations, most visibly on communication and forum networks with
	heterogeneous community structure: radoslaw-email ($+1.3$~pp), fb-forum ($+1.0$~pp), and
	enron-employees ($+0.4$~pp) at $d{=}100$. This matches the NC-LID analysis
	of~\cite{knevzevic2024local}: the datasets where NC-LID correlates most negatively with embedding
	quality under Dynnode2vec are those where the LID penalty helps most, since pairs with high
	NC-LID on both endpoints receive the largest loss amplification, preventing boundary nodes from
	collapsing into the interior of the nearest dominant community.
	
	On homogeneous proximity networks (primary-school, hospital), NC-LID scores are uniformly low,
	so the LID multiplier approaches 1 everywhere and the LID loss nears the baseline, explaining the
	near-parity there and confirming the signal is most valuable under structural ambiguity.
	
	\textbf{Distance\_Hub} wins zero of nine datasets and only 4 of 45 cells. The failure mode is the
	long-tailed degree-outer-product distribution on scale-free networks: a handful of extreme hub
	pairs dominate the gradient, sacrificing non-hub representations and distorting the embedding
	geometry, echoing~\cite{tomvcic2025dynamic}'s finding that hub-dominant walks hurt non-hub
	quality. NC-LID avoids this because its scores are bounded and roughly unimodal. Replacing
	$\delta_{ij}$ with $\log(1+\delta_{ij})$ would dampen the tails while preserving rank order, as
	in DeepHub's logarithmic scaling~\cite{tomvcic2025dynamic}.
	
	CollegeMsg and fb-messages, whose snapshots vary by up to a factor of 12 in node count, show a
	sharp $F_1$ drop at $d=200$ not seen elsewhere, plausibly because the one-hot feature matrix
	becomes extremely sparse for small snapshots when $N$ is large, poorly conditioning
	$\mathbf{W}^{(1)}\in\mathbb{R}^{N\times 2d}$ when $n_t \ll N$. No variant dominates both datasets
	at $d=100$: \textbf{Distance\_Basic} remains best on CollegeMsg (0.4641, vs.\ 0.4482 for
	\textbf{Distance\_LID}, $-1.59$~pp), while \textbf{Distance\_LID} leads on fb-messages (0.4471
	vs.\ 0.4320, $+1.51$~pp; Table~\ref{tab:res3}). This points to structural node features (degree,
	clustering coefficient) as a promising alternative to one-hot initialisation for highly
	variable-size graphs.

	\textit{Limitations.} Several aspects remain to be evaluated. (i) All variants share the Euclidean decoder, so the comparisons isolate the effect of the structural weighting but not of the decoder itself, a comparison with the standard inner-product GAE and with other distance-based decoders (e.g., cosine or hyperbolic) is needed to quantify its contribution. (ii) The sparsity correction is adopted from standard practice and was not ablated against the uncorrected BCE loss. (iii) Penalty strengths are fixed at $\alpha_h=\alpha_\ell=1$, a sensitivity analysis may reveal dataset-specific optima. (iv) Evaluation is limited to snapshot reconstruction, temporal tasks such as link prediction, other downstream tasks, and direct comparison with walk-based dynamic methods (Dynnode2vec, CTDNE, STWalk) are not covered. (v) On several datasets the margins between variants are small (below $0.5$~pp) and should be read as parity rather than as improvement.

	\section{Conclusions}\label{sec:conclusions}
	
	We introduced three distance-based graph autoencoder variants for dynamic graph embedding,
	sharing a Euclidean distance decoder that aligns the training objective with the evaluation
	metric. Augmenting the sparsity-corrected BCE loss with an NC-LID-based penalty, which amplifies
	errors for structurally ambiguous boundary nodes, improves reconstruction $F_1$ on most of the nine dynamic networks studied: \textbf{Distance\_LID} achieves the best $F_1$ in six of nine
	datasets and wins 26 of 45 dimension-dataset comparisons, confirming NC-LID, previously shown
	useful for walk-based methods~\cite{knevzevic2024local}, as an equally valuable signal within the
	autoencoder loss.
	
	\textbf{Distance\_Hub} consistently underperforms due to a gradient-concentration pathology from
	the long-tailed degree distribution, illustrating that structural penalties must be bounded to
	avoid distorting the embedding geometry, a requirement NC-LID satisfies naturally.
	
	Future work includes (i) ablations isolating the Euclidean decoder (against inner-product, cosine and hyperbolic decoders) and the sparsity correction, together with a sensitivity analysis of $\alpha_h$ and $\alpha_\ell$, (ii) temporally coupled training across snapshots and evaluation on link prediction and other downstream tasks, with comparisons to walk-based dynamic methods, (iii) logarithmic compression of the hub penalty to recover its motivation without the gradient pathology, (iv) replacing one-hot initialization with structural node features for highly variable-size graphs, and (v) extending the LID penalty to walk-based methods.
	
	\section*{Acknowledgements}
	This research is supported by the Science Fund of the Republic of Serbia, \#7462, Graphs in Space and Time: Graph Embeddings for Machine Learning in Complex Dynamical Systems -- TIGRA.
	 
	%
	%
	\bibliographystyle{spmpsci}
	\bibliography{refs}

@article{barros2021survey,
	title={A survey on embedding dynamic graphs},
	author={Barros, Claudio DT and Mendon{\c{c}}a, Matheus RF and Vieira, Alex B and Ziviani, Artur},
	journal={ACM Computing Surveys (CSUR)},
	volume={55},
	number={1},
	pages={1--37},
	year={2021},
	publisher={ACM New York, NY}
}

@inproceedings{grover2016node2vec,
	title={node2vec: Scalable feature learning for networks},
	author={Grover, Aditya and Leskovec, Jure},
	booktitle={Proceedings of the 22nd ACM SIGKDD international conference on Knowledge discovery and data mining},
	pages={855--864},
	year={2016}
}

@article{holme2012temporal,
	title={Temporal networks},
	author={Holme, Petter and Saram{\"a}ki, Jari},
	journal={Physics reports},
	volume={519},
	number={3},
	pages={97--125},
	year={2012},
	publisher={Elsevier}
}

@article{khoshraftar2024survey,
	title={A survey on graph representation learning methods},
	author={Khoshraftar, Shima and An, Aijun},
	journal={ACM Transactions on Intelligent Systems and Technology},
	volume={15},
	number={1},
	pages={1--55},
	year={2024},
	publisher={ACM New York, NY}
}

@article{kipf2016variational,
	title={Variational graph auto-encoders},
	author={Kipf, Thomas N and Welling, Max},
	journal={arXiv preprint arXiv:1611.07308},
	year={2016}
}

@article{kipf2016semi,
	title={Semi-supervised classification with graph convolutional networks},
	author={Kipf, Thomas N and Welling, Max},
	journal={arXiv preprint arXiv:1609.02907},
	year={2016}
}

@inproceedings{knevzevic2024local,
	title={Local intrinsic dimensionality for dynamic graph embeddings},
	author={Kne{\v{z}}evi{\'c}, Du{\v{s}}ica and Savi{\'c}, Milo{\v{s}} and Radovanovi{\'c}, Milo{\v{s}}},
	booktitle={International Conference on Complex Networks and Their Applications},
	pages={374--385},
	year={2024},
	organization={Springer}
}

@article{lancichinetti2009detecting,
	title={Detecting the overlapping and hierarchical community structure in complex networks},
	author={Lancichinetti, Andrea and Fortunato, Santo and Kert{\'e}sz, J{\'a}nos},
	journal={New journal of physics},
	volume={11},
	number={3},
	pages={033015},
	year={2009}
}

@inproceedings{mahdavi2018dynnode2vec,
	title={dynnode2vec: Scalable dynamic network embedding},
	author={Mahdavi, Sedigheh and Khoshraftar, Shima and An, Aijun},
	booktitle={2018 IEEE international conference on big data (Big Data)},
	pages={3762--3765},
	year={2018},
	organization={IEEE}
}

@inproceedings{nguyen2018continuous,
	title={Continuous-time dynamic network embeddings},
	author={Nguyen, Giang Hoang and Lee, John Boaz and Rossi, Ryan A and Ahmed, Nesreen K and Koh, Eunyee and Kim, Sungchul},
	booktitle={Companion proceedings of the the web conference 2018},
	pages={969--976},
	year={2018}
}

@article{nickel2017poincare,
	title={Poincar{\'e} embeddings for learning hierarchical representations},
	author={Nickel, Maximillian and Kiela, Douwe},
	journal={Advances in neural information processing systems},
	volume={30},
	year={2017}
}

@inproceedings{pandhre2018stwalk,
	title={Stwalk: learning trajectory representations in temporal graphs},
	author={Pandhre, Supriya and Mittal, Himangi and Gupta, Manish and Balasubramanian, Vineeth N},
	booktitle={Proceedings of the ACM India joint international conference on data science and management of data},
	pages={210--219},
	year={2018}
}

@inproceedings{perozzi2014deepwalk,
	title={Deepwalk: Online learning of social representations},
	author={Perozzi, Bryan and Al-Rfou, Rami and Skiena, Steven},
	booktitle={Proceedings of the 20th ACM SIGKDD international conference on Knowledge discovery and data mining},
	pages={701--710},
	year={2014}
}

@article{radovanovic2010hubs,
	title={Hubs in space: Popular nearest neighbors in high-dimensional data},
	author={Radovanovic, Milos and Nanopoulos, Alexandros and Ivanovic, Mirjana},
	journal={Journal of machine learning research},
	volume={11},
	number={sept},
	pages={2487--2531},
	year={2010}
}

@inproceedings{savic2021local,
	title={Local intrinsic dimensionality and graphs: towards LID-aware graph embedding algorithms},
	author={Savi{\'c}, Milo{\v{s}} and Kurbalija, Vladimir and Radovanovi{\'c}, Milo{\v{s}}},
	booktitle={International Conference on Similarity Search and Applications},
	pages={159--172},
	year={2021},
	organization={Springer}
}

@article{savic2023local,
	title={Local intrinsic dimensionality measures for graphs, with applications to graph embeddings},
	author={Savi{\'c}, Milo{\v{s}} and Kurbalija, Vladimir and Radovanovi{\'c}, Milo{\v{s}}},
	journal={Information Systems},
	volume={119},
	pages={102272},
	year={2023},
	publisher={Elsevier}
}

@article{tomvcic2024hub,
	title={Hub-aware random walk graph embedding methods for classification},
	author={Tom{\v{c}}i{\'c}, Aleksandar and Savi{\'c}, Milo{\v{s}} and Radovanovi{\'c}, Milo{\v{s}}},
	journal={Statistical Analysis and Data Mining: The ASA Data Science Journal},
	volume={17},
	number={2},
	pages={e11676},
	year={2024},
	publisher={Wiley Online Library}
}

@inproceedings{tomvcic2025dynamic,
	title={Dynamic graph embedding through hub-aware random walks},
	author={Tom{\v{c}}i{\'c}, Aleksandar and Savi{\'c}, Milo{\v{s}} and Simi{\'c}, Du{\v{s}}an and Radovanovi{\'c}, Milo{\v{s}}},
	booktitle={International Conference on Similarity Search and Applications},
	pages={315--329},
	year={2025},
	organization={Springer}
}

@article{zhang2018link,
	title={Link prediction based on graph neural networks},
	author={Zhang, Muhan and Chen, Yixin},
	journal={Advances in neural information processing systems},
	volume={31},
	year={2018}
}
	
\end{document}